\documentclass[]{style/ceurart}
\usepackage{booktabs}
\usepackage{multirow}
\usepackage{graphicx}
\usepackage{array}
\usepackage{caption}
\usepackage{listings}
\begin{document}

\copyrightyear{2024}
\copyrightclause{Copyright for this paper by its authors.
  Use permitted under Creative Commons License Attribution 4.0
  International (CC BY 4.0).}

\conference{CLEF 2024: Conference and Labs of the Evaluation Forum, September 9-12, 2024, Grenoble, France}

\title{
DS@GT eRisk 2024: Sentence Transformers for Social Media Risk Assessment
}

\author[1]{David Guecha}[
email=dahumada3@gatech.edu,
orcid=0009-0009-9855-5330
]
\cormark[1]
\author[1]{Aaryan Potdar}[
email=apotdar31@gatech.edu
]
\author[1]{Anthony Miyaguchi}[
orcid=0000-0002-9165-8718,
email=acmiyaguchi@gatech.edu,
]

\address[1]{Georgia Institute of Technology, North Ave NW, Atlanta, GA 30332}
\cortext[1]{Corresponding author.}

\begin{abstract}
We present working notes for DS@GT team in the eRisk 2024 for Tasks 1 and 3. 
We propose a ranking system for Task 1 that predicts symptoms of depression based on the Beck Depression Inventory (BDI-II) questionnaire using binary classifiers trained on question relevancy as a proxy for ranking.
We find that binary classifiers are not well calibrated for ranking, and perform poorly during evaluation.
For Task 3, we use embeddings from BERT to predict the severity of eating disorder symptoms based on user post history.
We find that classical machine learning models perform well on the task, and end up competitive with the baseline models.
Representation of text data is crucial in both tasks, and we find that sentence transformers are a powerful tool for downstream modeling.
Source code and models are available at \url{https://github.com/dsgt-kaggle-clef/erisk-2024}.

\end{abstract}


\maketitle

\section{Introduction}

The eRisk Challenge 2024 \cite{parapar_overview_2024, parapar_wn_overview_2024} is comprised of three distinct tasks aimed at developing early risk prediction systems that utilize social media documents to detect antisocial behaviors, signs of mental illnesses, and eating disorders.
Users often view social media as an ideal platform for self-expression. They can share personal experiences and anecdotes while remaining anonymous to those on the platform.
The openness of social media platforms allows users to seek support and advice from others without the fear of being judged.
These factors make social media, such as Reddit, a valuable resource for mental health research.
By analyzing the content shared by users, we can potentially identify early signs of eating disorders or symptoms of depression and assess their severity.

Our team focused on two specific tasks: task 1, which involved identifying depression symptoms from a questionnaire, and task 3, which focused on diagnosing eating disorders.
For task 1, we proposed a system that predicts symptoms of depression based on the Beck Depression Inventory (BDI-II)\cite{beck_beck_1996} using various sentence processing architectures and a multi-label classification approach.
For task 3, we developed a system that predicts symptoms of eating disorders by learning classical machine-learning models using vector representations of user posts retrieved from sentence transformers.

\subsection{Related Work}

The eRisk challenge, as detailed in \citet{crestani_early_2022}, has a rich history of over seven years of experiments to draw from.
In Task 1 of the 2023 challenge\cite{parapar_overview_2023}, various teams employed vector representations of sentences and leveraged either semantic search with transformer-based models or cosine similarity to categorize documents according to the 21 symptoms of depression \cite{recharla_notebook_nodate,wang_notebook_nodate}.
For Task 3, teams such as those reported by \citet{grigore_notebook_nodate} experimented with domain-specific language models, including MentalBERT, and utilized topic modeling to make their predictions.

\section{Task 1: Search for Symptoms of Depression}

Task 1 involves ranking documents relevant to symptoms of depression as identified in the BDI-II questionnaire.
The objective is to submit runs containing the top thousand documents pertinent to a specific symptom.
This is a standard information retrieval task commonly applied in search engines.
The evaluation is conducted against a pool of human assessors who are experts in the field, categorizing sentences as either relevant or not relevant.
There are two types of question relevancey (qrel) scoring: majority and unanimity \cite{noauthor_text_nodate}.
Human-expert question relevancy scores were made available in this year's competition, allowing these documents to be used as training data.

Inspired by last year's eRisk competition participants, particularly the Formula-ML team who achieved the highest scores across the board, we chose to build our model using sentence transformers alongside traditional NLP methods. Our goal was to compare and contrast the performance of count methods, vector models, and rich semantic representations on early risk detection systems.

This task is well-suited for supervised learning approaches. We utilized logistic regression to make binary relevance predictions for each symptom in the questionnaire, optimizing for F1 scores and accuracy using the relevance labels from the QRELS provided by the lab as our baseline for binary relevance decisions. Our internal validation showed that different iterations of our models achieved over 70\% in mean accuracy and mean F1 scores, with sentence transformers achieving the highest scores of 89\% mean F1 and 90\% mean accuracy. This gave us confidence in submitting our results using the sentence transformers and Word2Vec models.

\subsection{Dataset}

A training and test TREC dataset contains user posts from Reddit.
It contains over 500k users with almost 20 million sentences.

\begin{table}[h]
  \caption{Exploratory statistics for the Task 1 dataset.}
  \label{tab:task1-stats}
  \centering
  \begin{tabular}{lrr}
      \toprule
       & \textbf{Test Set} & \textbf{Training Set} \\
      \midrule
      \textbf{Number of Users} & 552,315 & 3,107 \\
      \textbf{Number of Sentences} & 15,542,200 & 4,264,693 \\
      \textbf{Average number of words per sentence} & 17.96 & 13.95 \\
      \textbf{Median number of words per sentence} & 16 & 11 \\
      \bottomrule
  \end{tabular}
\end{table}

The dataset was formatted as TREC documents, each containing a unique document number (DOCNO) and a text field (TEXT) containing the post content.
The test files contain additional fields PRE and POST fields with text content that may add context to the TEXT field.
We compute some statistics on the dataset, including the number of users, the number of sentences, and the average and median number of words per sentence in Table \ref{tab:task1-stats}.

\begin{figure}[h]
\centering
\begin{lstlisting}[language=xml,frame=single]<DOC>
<DOCNO>s_0_2_4</DOCNO>
    <TEXT>I feel depressed</TEXT>
</DOC>
\end{lstlisting}
\caption{
  Example post in a TREC document.
  The DOCNO field contains the document number, and the TEXT field contains the post content.
}
\label{fig:example-trec}
\end{figure}

\subsection{Methodology}

Two distinct modeling approaches were employed for Task 1: a baseline and a sentence-transformer-based solution.
The baseline system uses classical classifiction models (e.g. Naive Bayes) trained on binary relevance labels as a proxy for relevance ranking.
The sentence-transformer-based model does the same, but uses sentence embeddings as the underlying representation of the text.

We ran our experiments on Google Cloud Platform (GCP) compute instance sized n1-standard-2.
We used PySpark \cite{noauthor_pyspark_nodate} for parallel processing and handling large data volumes.
Luigi \cite{luigi} was used to define the data processing pipelines and workflows.


\begin{figure}[h]
  \centering
  \includegraphics[width=\textwidth]{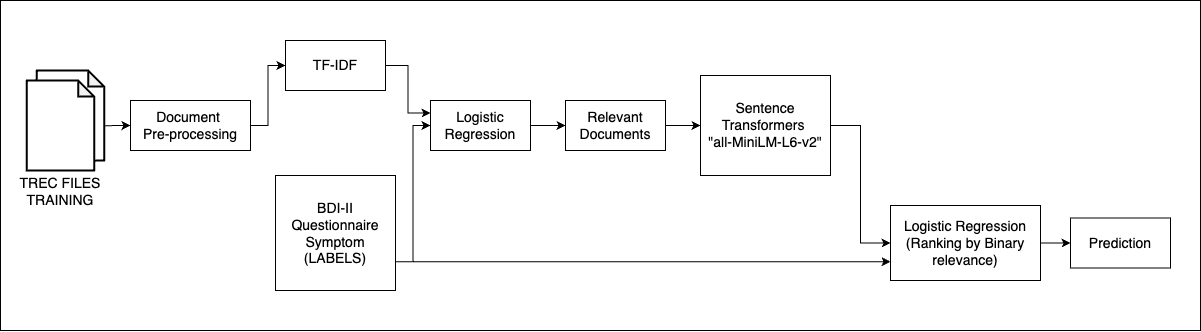}
  \caption{
    The modeling pipeline for Task 1 using sentence transformers.
    Binary relevance labels train a classifier that ranks documents based on their relevance to the BDI-II questionnaire.
    These relevance predictions help filter documents to limit transformation computation with a sentence transformer model.
    The final model ranks the documents based on their relevance to the BDI-II questionnaire.
  }
  \label{fig:pipeline}
\end{figure}  

\subsubsection{Preprocessing}

We collected the TREC formatted files and undertook extensive data cleansing, including the removal of special characters and the correction of formatting errors.
Subsequently, we merged the cleaned files using a python script that compressed the documents into the Parquet format, a binary format optimized for fast querying and favored over CSV and other formats.
To ensure both persistence and easy access during model development, we hosted the data in Google Cloud Buckets, and access to the data was facilitated through the Google Cloud API.

\subsubsection{Modeling}

We trained a classifier for each question of the BDI-II to determine document relevancy to specific questions.
Each document was evaluated against the classifier to obtain a relevancy probability, which was then used to rank the documents.
The training data was vectorized using a count vectorizer, Word2Vec, and a text transformer.
We hypothesized that framing the ranking task as a classification problem would yield satisfactory performance on the leaderboard.
Additionally, we posited that unsupervised text representations with higher learning capacities would perform better in the relevancy classification task.
See Table \ref{tab:task1-models} for a summary of the modeling approaches.

\begin{table}[h]
  \centering
  \caption{
    Modeling approaches used for Task 1.
    Binary classification models are trained on the BDI-II questionnaire to rank documents based on their relevance to the questionnaire.
  }

  \label{tab:task1-models}
  \renewcommand{\arraystretch}{1.2}
  \begin{tabular}{p{2.5in}p{3.5in}}
  \toprule
  \textbf{Model} & \textbf{Description} \\ \midrule
  Naive Bayes - Counting Vectorizer & Discriminator based on feature-independence of words. Only usable with positive features. \\ \midrule
  Logistic Regression - Counting Vectorizer & Classifier using vector of word counts, learns coefficients that determine how to weight words to fit a decision boundary. \\ \midrule
  Logistic Regression - Word2Vec  & Classifier using a word embedding that captures distributional semantics of the bag of words model. The unsupervised task should transfer knowledge to the simpler linear classifier. \\ \midrule
  Logistic Regression - Text Transformer & Classifier using the inductive properties of transformer layers to model a sequence of words in an auto-regressive manner. \\ \bottomrule
  \end{tabular}
\end{table}

We transform text from it's natural representation into a representation that can be used by a machine learning model.
The counting vectorizer treats each document as a vector by counting the frequency of each word in the document.
TF-IDF is a normalization of the count vector that accounts for the term-frequency (TF) and inverse document frequency (IDF) of each word.
Word2Vec is a neural network model that learns word embeddings by predicting the context of a word in a sentence, using either the continuous bag of words (CBOW) or skip-gram model.

We also used the sentence transformer model \cite{reimers_sentence-bert_2019} to generate sentence embeddings.
The model we utilized for getting the embeddings was \texttt{all-MiniLM-L6-v2} that transforms text into a vector in $\mathbb{R}^{384}$.
The language model is designed to be general purpose and relatively fast for natural language tasks \cite{noauthor_sentence-transformersall-minilm-l6-v2_nodate}.
These models are trained on large corpus of text data, and learn to encode the meaning of a sentence into a fixed-length vector through sequence alignment and attention mechanisms.
In order to use sentence transformers with Spark, we implemented a user-defined function (UDF) wrapped in a Python class to generate the embeddings from the TEXT column.

We had to aggressively filter out low-quality sentences since embedding documents took many hours for hundreds of thousands of documents out of millions.
We used relevancy predictions from a simpler logistic regression model on count vectorized data to filter out irrelevant documents before transforming them with the sentence transformer model.
We also experimented with filtering based on the compression ratio of the text, as we found both low and high compression ratios to be indicative of irrelevant documents.
The compression ratio is the ratio of the size of gzip compressed text to the size of the original text.
The mean compression ratio of the documents was 0.9 with a standard deviation of 0.1.

\subsection{Results}

We report the results of our models on the public leaderboard in Table \ref{tab:task1-results}.
After evaluating the hidden test set, our models score zero nearly across the board.
We find that the transformer-based model scores an order of magnitude more than the models without the recall-precision and NDCG scores.

\begin{table}[h]
\centering
\caption{Ranking-based evaluation for Task 1}
\label{tab:task1-results}
\begin{tabular}{lllllllllll}
\toprule
Run & \multicolumn{4}{c}{Unanimity} & & \multicolumn{4}{c}{Majority Voting} \\
\cmidrule{2-5} \cmidrule{7-10}
 & MAP & R-PREC & P@10 & NDCG & & MAP & R-PREC & P@10 & NDCG \\
\midrule
logistic\_transformer\_v5 & 0.000 & 0.006 & 0.000 & 0.010 & & 0.000 & 0.009 & 0.000 & 0.014 \\
logistic\_word2vec\_v5 & 0.000 & 0.001 & 0.000 & 0.003 & & 0.000 & 0.001 & 0.000 & 0.003 \\
count\_logistic & 0.000 & 0.000 & 0.000 & 0.001 & & 0.000 & 0.000 & 0.000 & 0.001 \\
count\_nb & 0.000 & 0.000 & 0.000 & 0.000 & & 0.000 & 0.000 & 0.000 & 0.000 \\
word2vec\_logistic & 0.000 & 0.000 & 0.000 & 0.000 & & 0.000 & 0.000 & 0.000 & 0.000 \\
\bottomrule
\end{tabular}
\end{table}

\subsection{Discussion and Future Work}

In this project, we assumed that a classifier could be repurposed as a functional ranker.
However, our results on the leaderboard contradicted this assumption.
One potential reason for the poor performance is that a classifier optimized for a specific metric (such as F1 or accuracy) may not be well-calibrated to the actual data distribution.
Given more time, we would have explored proper ordinal regression through learning to rank methodologies.

During pipeline construction, we observed a non-trivial number of training examples with high repetition.
We would have preferred to explore more options to reduce the number of relevant training examples.
One potential solution to eliminate irrelevant documents would be to filter relevance using keyword-based information retrieval algorithms \cite{robertson_probabilistic_2009}, before ranking the documents with a statistical model.
These repetitive documents indicated a poorly performing model, as they represented ill-behaved, degenerate examples that the model struggled to capture due to representation issues.
For instance, a word2vec model trained on repeating fragments like "I AM SAD" would likely place these fragments in the same semantic space as a single instance, making differentiation based on angle alone tricky.
Thus, sentences with high compression ratios and low entropy should be filtered from the dataset.
Our qualitative evaluation during development indicated that filtering out such sentences improved retrieval performance.

For future iterations of this task, we propose leveraging various language models to explore the capabilities of transformer-based systems.
Additional strategies could include employing prompt engineering, given that large language models are effective search engines; this could prove to be a compelling area of exploration for information retrieval systems.
Other potential avenues include improving our pre-filtering process to remove non-relevant posts, fine-tuning pre-trained language models, and employing retrieval-augmented generation(RAG) to enhance the accuracy of predictions.

\section{Task 3: Measuring the Severity of the Signs of Eating Disorders}

Eating disorders (ED) are serious mental health conditions characterized by abnormal eating habits.
Early detection and severity assessment are crucial for timely intervention and treatment.
Task 3 aims to explore the feasibility of automatically estimating the severity of symptoms of ED using the Eating Disorder Examination Questionnaire (EDE-Q) based on the activity of social media users.
We aimed to leverage state-of-the-art NLP techniques and Machine Learning methods to design a pipeline for detecting the severity of the signs of EDs.
Our approach of BERT-based \cite{devlin_bert_2019} text embeddings enabled us to perform well with limited available information.

\subsection{Dataset}

For this task, participants developed a system for predicting responses to an Eating Disorder Examination Questionnaire based on a Reddit post history.
The EDE questionnaire is a 28-item self-reported questionnaire adapted from the Eating Disorder Examination (EDE).
The questionnaire covers several domains: dietary restraint, eating, shape, and weight concerns.
The responses range from 0 to 6, corresponding to the severity of symptoms.
Our goal was to predict the responses to 22 of 28 EDE-Q questions based on user post history.

\begin{table}[h!]
\centering
\caption{Exploratory statistics for the Task 3 dataset.}
\begin{tabular}{|c|l|r|}
\hline
\multirow{4}{*}{2022} & No. of Subjects                 & 28     \\ \cline{2-3}
                      & Min. no. of posts per Subject   & 12     \\ \cline{2-3}
                      & Max. no. of posts per Subject   & 1143   \\ \cline{2-3}
                      & Avg. no. of characters per Post & 184.33 \\ \hline
\multirow{4}{*}{2023} & No. of Subjects                 & 46     \\ \cline{2-3}
                      & Min. no. of posts per Subject   & 5      \\ \cline{2-3}
                      & Max. no. of posts per Subject   & 1161   \\ \cline{2-3}
                      & Avg. no. of characters per Post & 223.25 \\ \hline
\end{tabular}
\end{table}

During the training phase of the challenge, the eRisk team released an entire history of user writings along with the corresponding answers to the EDE Questionnaire responses.
The training data includes users from the 2022 and 2023 datasets, while the test data comprises new users from 2023.
The combined training set consisted of 74 subjects.
For each user or subject, we had access to a history of postings on Reddit and their responses to the EDE-Q questionnaire.
We used the user responses as ground truth for training our machine learning models (discussed in the methodology section).
The test set consisted of a history of postings for 18 users, structured similarly to the training set.

\subsection{Methodology}

\begin{figure}[h]
  \centering
  \includegraphics[width=\textwidth]{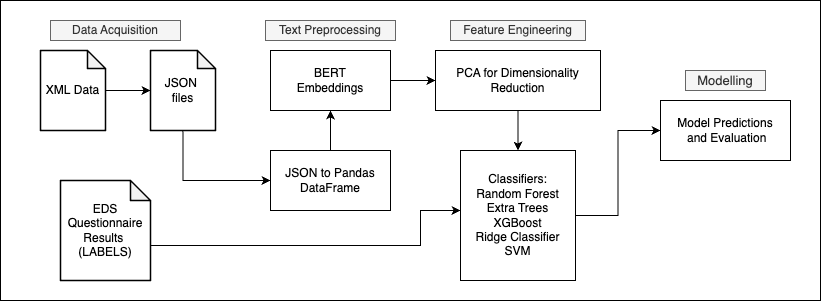}
  \caption{
    A diagram of the Task 3 pipeline.
    Each user's post history is fed into a BERT model to generate embeddings.
    The embeddings are then fed into a machine-learning model to predict the EDE-Q responses.
  }

  \label{fig:example}
\end{figure}

The goal of task 3 was to predict the responses of the 2024 users to the EDE-Q questionnaire.
To achieve this, we needed to determine to what extent the characteristics associated with eating disorders are reflected in the social media user's post and comment history.
Note that the responses to the EDE-Q are integers ranging from 0 to 6.
We approach this problem as a multi-label and multi-output classification task.

\subsubsection{Preprocessing}

The eRisk lab organizers made datasets available for the 2022 and 2023 users.
We processed TREC data into JSON using BeautifulSoup and \texttt{etree.XMLParser}.
Preprocessing involved converting the JSON files to readable DataFrames in Pandas and cleaning the text data to remove noise, such as URLs, hashtags, and special characters.
We tokenized the text and applied lemmatization, stemming, and stopword removal to normalize the data.

We leveraged BERT to generate text embeddings.
For this task, we used the \texttt{bert-base-uncased} \cite{noauthor_google-bertbert-base-uncased_nodate} pre-trained language model, which has been trained on 110 million parameters and works well with English texts. The preprocessing times for generating BERT embeddings are summarized below.

\begin{table}[h]
\centering
\caption{Preprocessing Times for BERT Embeddings}
\begin{tabular}{|c|c|c|}
\hline
\textbf{Dataset} & \textbf{Time (minutes:seconds)} & \textbf{Number of Chunks} \\ \hline
2022 Dataset     & 21:11                          & 8477                      \\ \hline
2023 Dataset     & 24:39                          & 15133                     \\ \hline
2024 Test Dataset& 18:22                          & 7519                      \\ \hline
\end{tabular}
\label{table:preprocessing_times}
\end{table}

BERT has a maximum input sentence length of 512 tokens, including the [CLS] and [SEP] tokens, and generates vectorized embeddings in $\mathbb{R}^{768}$.
For posts that were too long to feed into the transformer model, we concatenated all user posts together in chronological order. We broke the resulting text into chunks of length $n$, where $n = 512 - 2 = 510$.

We hypothesized that the 768 dimensions of the sentence embeddings might be too large.
Therefore, we standardized the characteristics and performed principal component analysis (PCA) for dimensionality reduction \cite{fodor_survey_2002} to reduce the dimension down to 50.
In the model training phase, we trained the machine learning models on both high and low dimensional embeddings to compare the performance of the models.

\subsubsection{Modeling}

Given the substantial data requirements of deep learning models, we opted for classical machine learning models for our analysis.
We selected five models: Random Forest, Extra Trees, XGBoost, Ridge Regression, and Support Vector Machines (SVM).
Random Forest and Extra Trees are ensemble methods that combine multiple decision trees to improve performance and reduce overfitting.
XGBoost is a gradient boosting algorithm that is known for its speed and performance on tabular data.
Ridge Regression is a linear model that incorporates L2 regularization to prevent overfitting.
SVM is effective for high-dimensional data and can handle non-linear relationships.
For a model baseline, we used results from the runs for the 2023 challenge, which involved the same task.

To evaluate the performance of the models, the eRisk team provided several evaluation metrics: Mean Zero-One Error, Mean Absolute Error, Macro-averaged Mean Absolute Error, Restraint Subscale, Eating Concern Subscale, Shape Concern Subscale, Weight Concern Subscale, and Global ED (the global score).

\subsection{Results}

\begin{figure}[h]
  \centering
  \includegraphics[width=0.8\textwidth]{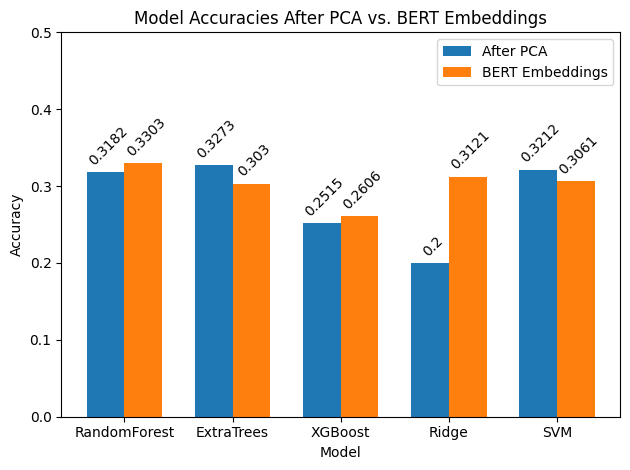}
  \caption{Task 3 Model performance on vector embedding with high dimensions and after dimensionality reduction.}
  \label{fig:example}
\end{figure}

Random Forest Classifier with high-dimensional embeddings performed the best overall, achieving an accuracy of 0.3303 and an MAE of 2.1091.
This model benefited from the high-dimensional feature space, which likely captured the complex patterns in the data more effectively.
After applying dimensionality reduction, the Extra Trees Classifier emerged as the top performer with an accuracy of 0.3273.
This suggests that reducing the feature dimensions helped improve the model's generalizability.
Comparatively, the XGBoost Classifier showed lower performance across both feature sets, with an accuracy of 0.2606 and 0.2515 for high-dimensional and reduced-dimensional data, respectively.
This may be due to the model's sensitivity to hyperparameters, which requires careful optimization.

The Ridge Classifier and SVM showed moderate performance.
Notably, the Ridge Classifier's performance significantly dropped when trained on reduced dimensions data, with an accuracy of 0.
2000, highlighting its limitation in handling reduced feature spaces effectively.
The SVM performed consistently across both feature sets, but its overall accuracy remained lower compared to Random Forest and Extra Trees.

\begin{table}[h]
  \caption{Results for Task 3 on the public leaderboard.}
  \label{tab:task3-results}
  \begin{tabular}{llllllllll}
  \toprule
  team     & run ID  & MAE   & MZOE  & MAEmacro & GED   & RS    & ECS   & SCS   & WCS   \\ \midrule
  baseline & all 0s  & 3.79  & 0.813 & 4.254    & 4.472 & 3.869 & 4.479 & 4.363 & 3.361 \\
  baseline & all 6s  & 1.937 & 0.551 & 3.018    & 3.076 & 3.352 & 2.868 & 3.029 & 2.472 \\
  baseline & average & 1.965 & 0.884 & 1.973    & 2.337 & 2.486 & 1.559 & 2.002 & 1.783 \\ \midrule
  DSGT     & 0       & 1.965 & 0.588 & 1.713    & 2.211 & 2.321 & 1.969 & 1.944 & 2.117 \\ \bottomrule
  \end{tabular}
\end{table}

We report the results of our models on the public leaderboard in Table \ref{tab:task3-results}.
While our models generally met baseline metrics, they significantly outperformed the baseline in Mean Zero-One Error (MZOE), Mean Absolute Error (MAE), and MAE macro metrics on the test data. 
However, our models fell short in meeting the baseline for the Eating Concern (ECS) and Weight Concern (WCS) subscales, suggesting a need for additional techniques like topic modeling or semantic analysis.
Despite these challenges, our team consistently ranked in the top 5 among five participating teams, with a total of 14 submissions. This highlights the robustness of our approach.


\subsection{Discussion and Future Work}

Our study achieved promising results in predicting the severity of eating disorder symptoms from social media content. 
Further research should focus on refining our models and incorporating advanced techniques to enhance performance, ultimately contributing to improved interventions and support systems for individuals with eating disorders.

Further research may explore the potential of deep learning models for predicting the severity of eating disorder symptoms from social media content. 
Deep learning architectures, such as recurrent neural networks (RNNs), offer the ability to capture complex patterns in text data more effectively, which could lead to improved performance.
Given that deep learning models often require large amounts of data, we would need to explore data augmentation techniques. 
We could generate additional data using techniques such as Language Model Fine-Tuning (LMFT) or services like TextSynth to increase the diversity and size of our dataset.


\section{Conclusion}

In this document, we presented our working notes for the eRisk lab of the CLEF 2024 conference, where we submitted entries for Task 1 and Task 3.
In Task 1, we propose a system that ranks documents based on their relevance to the BDI-II questionnaire using a binary classifier as a proxy for ranking.
The performance is lackluster, and future work would explore proper ordinal regression methods such as learning to rank.
In Task 3, we developed a system that predicts the severity of eating disorder symptoms based on user post history.
We use BERT embeddings to generate features for classical machine learning models, and prove competitive with the baseline models.

In all our tasks, we find that sentence transformers are a powerful tool in representing text data, and that the choice of model and feature representation can significantly impact the performance of the system.
Source code and models are available at \url{https://github.com/dsgt-kaggle-clef/erisk-2024}.

\section*{Acknowledgements}

Thank you to the DS@GT CLEF for subsidizing compute and storage used for this project.
Thank you to the eRisk organizers for providing the dataset and evaluation for the competition.

\bibliography{DSGT-eRisk}


\end{document}